%% file: acl.tex
\pdfoutput=1

\documentclass[11pt]{article}

\usepackage{acl}

\usepackage{times}
\usepackage{latexsym}

\usepackage[T1]{fontenc}

\usepackage[utf8]{inputenc}

\usepackage{microtype}

\usepackage{colortbl}
\usepackage{times}
\usepackage{latexsym}

\usepackage{booktabs}
\usepackage{subfigure}
\usepackage{graphicx}
\usepackage{multirow}
\usepackage{float} 
\usepackage{color}
\usepackage{xcolor}
\usepackage{url}
\usepackage{wrapfig}
\usepackage{setspace}
\usepackage{amssymb}
\usepackage{pifont}
\usepackage{appendix}
\usepackage{algorithm}
\usepackage{algorithmic}
\usepackage{amsmath}
\usepackage{CJKutf8}
\usepackage{pifont}
\usepackage{bm}
\usepackage{tabularx}
\usepackage{ulem}
\usepackage{longtable}
\usepackage{pdfpages}

\newcommand{\ie}{\textit{i.e.}}
\newcommand{\eg}{\textit{e.g.}}

\newcommand{\correspondingauthor}{\thanks{~~Corresponding authors.}}

\title{Reverse Modeling in Large Language Models}

\author{
   Sicheng Yu$^{1}$\thanks{$\;\;$The first two authors contributed equally to this work.} 
   ~~Yuanchen Xu$^{2*}$
   ~~Cunxiao Du$^{1}$
   ~~Yanying Zhou$^{3}$ \\ \bf
   Minghui Qiu$^{1}$
   ~~Qianru Sun$^{1}$
   ~~Hao Zhang$^{4}$\correspondingauthor
    ~~~~Jiawei Wu$^{2}$\footnotemark[2]
   \\
   $^{1}$Singapore Management University ~~
   $^{2}$National University of Singapore \\
   $^{3}$Fudan University ~~
   $^{4}$DAMO Academy, Alibaba Group \\
\texttt{scyu.2018@phdcs.smu.edu.sg}~~\\
\texttt{\{yuanchen\_xu,jiaweiwu\}@u.nus.edu}~~\\
   \texttt{hz.hhea2e@alibaba-inc.com}~~
}

\begin{document}
\maketitle

\input{sections/0_abs}
\input{sections/1_intro}
\input{sections/2_related_work}
\input{sections/3_exp_setting}

\input{sections/4_exp_res}
\input{sections/5_con}
\input{sections/6_limitation}

\section*{Acknowledgments}
We would like to express our sincere gratitude to the anonymous reviewers for their insightful comments and valuable suggestions, which substantially improved this manuscript. 
S. Y. and J. W. are indebted to Xiaoqun Xiao, whose expertise in Arabic linguistics and textual analysis is instrumental in developing the multilingual aspects of this work.

\bibliography{anthology}
\bibliographystyle{acl_natbib}


\input{sections/7_appendix}
\end{document}

%% file: sections/0_abs.tex
\begin{abstract}
Humans are accustomed to reading and writing in a forward manner, and this natural bias extends to text understanding in auto-regressive large language models (LLMs).
This paper investigates whether LLMs, like humans, struggle with reverse modeling, specifically with reversed text inputs.
We found that publicly available pre-trained LLMs cannot understand such inputs. However, LLMs trained from scratch with both forward and reverse texts can understand them equally well during inference across multiple languages.
Our case study shows that different-content texts result in different losses if input (to LLMs) in different directions---some get lower losses for forward while some for reverse.
This leads us to a simple and nice solution for data selection based on the loss differences between forward and reverse directions.
Using our selected data in continued pretraining can boost LLMs' performance by a large margin across different language understanding benchmarks.
\end{abstract}

%% file: sections/1_intro.tex
\section{Introduction}
\label{sec:intro}

LLMs~\cite{touvron2023llama,jiang2023mistral} 
have shown impressive capabilities in various natural language processing tasks and beyond.
These capabilities are primarily attributed to the learning of extensive corpora that cover general world knowledge~\cite{kaplan2020scaling}.
These
corpora are created in human society and often demonstrate human bias, including inherently forward-oriented human cognition~\cite{bergen2005writing,de2013alphabet}, 
\eg, reasons may precede outcomes and solutions can be deduced from given information in most cases of the grad school math dataset~\cite{mitra2024orca}.
In contrast, reverse thinking presents more cognitive challenges due to its contradiction with innate common sense and human logic~\cite{chen2024reverse}.
It inspires us to explore the following questions:
\begin{itemize}
    \item \textit{Can LLMs perform reverse modeling or will they face similar challenges as humans?}
    \item \textit{Can reverse modeling benefit the learning of LLMs?}
\end{itemize}

To study this, we simulate reverse-modeling data by directly reversing entire paragraphs or documents at the token level.
Please note that this is the simplest and extreme way, but may not be the optimal way of emulating reverse thinking.
We train LLMs with these simulated texts and conduct a comprehensive analysis. 
Overall results indicate that LLMs learn forward- and reverse-modeling texts equally well when trained from scratch. 
However, performance varies across text samples. Some are suited to reverse modeling, while others favor forward modeling.
Notably, we find that the texts suited for reverse modeling are often of high quality and more logically coherent. Training on them, the original ``forward-modeling'' LLMs can be improved.
We perform empirical validation on language understanding benchmarks, such as Massive Multitask Language Understanding (MMLU)~\cite{hendrycks2020measuring}.
In summary, this paper has two main contributions.
\begin{itemize}
\item We examine how LLMs process and learn from text in both forward and reverse directions, demonstrating consistent patterns across multiple languages.
\item We show that strategically selecting training data based on the losses of forward- and reverse-modeling leads to improved model capabilities.
\end{itemize} 

%% file: sections/2_related_work.tex
\section{Related Work}
In this paper we utilize the reverse text for model training.
Previous work on reverse inputs falls into three main areas. 
The first area involves the use of reverse text in machine translation. Studies show that using decoders to process text both left-to-right and right-to-left within an encoder-decoder framework improves machine translation performance~\cite{zhou2019synchronous,gu2019insertion}, a finding later extended to LLMs~\cite{nguyen2024meet}. Concurrently, \cite{wu2018beyond} examines the relationship between error propagation and reverse direction decoding in machine translation. 
The second area focuses on the reversal curse~\cite{berglund2023reversal,zhu2024towards}, where an LLM trained to understand ``A is B'' may struggle to generalize to ``B is A''. Reversing the text is proposed as a solution to this problem~\cite{golovneva2024reverse,guo2024mitigating}. 
These two streams of work focus on machine translation or the reversal curse.
Third, a recent work~\cite{papadopoulos2024arrows} also explores the direction of input text, but there are two key differences compared to ours: (1) Our work is inspired by the concept of reverse thinking, while the reversed input is one simulating solution; (2) We further analyze it across different domains and inference steps and discover a valuable tool for assessing data quality.

Our applications are partially related to the selection of training data for LLMs, which is divided mainly into heuristic and model-based methods~\cite{longpre2023pretrainer}. 
Heuristic methods filter out low-quality data by defining various rules, such as the ratio of nouns and verbs~\cite{raffel2020exploring, penedo2023refinedweb, chowdhery2023palm, sharma2024text}. Model-based methods filter data by training selection models or based on the perplexity of language models~\cite{wenzek2019ccnet, xie2023data, wettig2024qurating}. 
However, our data selection method is an extra bonus derived from the reverse modeling analysis.

%% file: sections/3_exp_setting.tex
\section{Experimental Settings}

\paragraph{Forward and Reverse Training.} Given a original text, it can be represented as a sequence
after tokenization, which is used for forward training.
To perform reverse training, we directly reverse the original token sequence
to construct a reverse training sample. While some studies explore keeping the original orders of detected words or entities during reverse~\cite{golovneva2024reverse,guo2024mitigating}, we choose the simplest operation to avoid the various performance of detection modules in different domains and languages. 
The Llama2-7B~\cite{touvron2023llama} (or the randomly initialized version) is selected as the default backbone in this paper.

\paragraph{Datasets.}
In Research Question (RQ) 1, we used the multilingual mC4\footnote{English, German, Korean, Arabic from \url{https://huggingface.co/datasets/allenai/c4}}~\cite{raffel2020exploring} dataset to compare LLMs' ability to handle forward and reverse texts under continued and from-scratch pretraining settings. In subsequent experiments, we used the English SlimPajama\footnote{We use the widely-used public sampled version for experiments: \url{https://huggingface.co/datasets/DKYoon/SlimPajama-6B}}~\cite{soboleva2023slimpajama} dataset, which includes seven different source domains. Testing LLMs trained on the multilingual mC4 dataset with samples from the SlimPajama dataset can further confirm our findings are general. More details are in Appendix~\ref{app:implement}.

%% file: sections/4_exp_res.tex
\section{Experiments}
\label{section:exp}
\subsection*{RQ1: \textit{Can LLMs perform reverse modeling?}}

\input{figures/multilingual_loss_en}

\input{tables/main_cases}

To explore LLMs' reverse modeling capabilities, we investigate two pre-training approaches: (1) continued training from a well-trained model checkpoint and (2) pretraining from scratch with random initialization. 
Specifically, we train models fed with forward input and reverse text using the two approaches separately.
Figure~\ref{figure:multilingual_loss_en} compares training losses (average sample losses within training batches) for English using both methods on the mC4 dataset, while Figure~\ref{figure:multilingual_loss} in the Appendix~\ref{app:multilingual} shows analogous results for other languages.

In the continued pretraining setting, the forward loss for forward-modeling remains stable due to extensive training in the initial pretraining stage. In contrast, the reverse loss for reverse modeling, initially high, decreases rapidly after a few training steps. 
Notably, the forward loss is consistently lower than the reverse loss during continued pretraining. We speculate this occurs because the initial pretraining corpora consists entirely of forward-direction texts, imparting a natural directional bias to the LLMs. Consequently, the models find processing reverse information more challenging, similar to human difficulties with reverse thinking.

Interestingly, in the from-scratch pretraining, the loss curves for both text directions converge almost identically. This pattern, also observed in other languages, indicates that LLMs can learn to process forward and reverse-modeling inputs with similar proficiency. This is because the model learns from both forward and reverse texts simultaneously with randomly initialized parameters, avoiding the initial forward-direction bias in well-trained models.

\input{figures/slim_diff_loss_domain_distribution}

\subsection*{RQ2: \textit{Does data domain influence the ability of LLMs' reverse modeling?}}

Based on the observation in RQ1, we focus on the from-scratch pretraining setting, where trained LLMs show almost equal losses from both forward and reverse directions. 
This raises the question of whether reverse loss consistently equals forward loss across all texts or if there are instances where reverse learning incurs a lower or higher loss. 
To explore this, we use the SlimPajama~\cite{soboleva2023slimpajama} text dataset, which covers a broad range of domains, for case-level evaluation.

Given a text sequence represented by tokens $\{V_1, V_2, \cdots, V_N\}$, for each position $t$ in the sequence ($0 \leq t \leq N$), a LLM can generate a probability distribution over possible next tokens. We compute the cross-entropy loss at each position $t$, resulting in two sequences of loss values: $\{F_1, F_2, \cdots, F_N\}$ for the forward sequence and $\{R_1, R_2, \cdots, R_N\}$ for the reverse sequence.

We first compute the average loss difference (computed as $\frac{1}{N}(\sum_{i=1}^N F_i - \sum_{i=1}^N N_i)$) for each text and associate each text with its corresponding data source label.
The overall case-level loss difference distribution across different source domains is shown in Figure~\ref{figure:diff_domain}. 
Observed that the loss differences of the text samples are centered around zero, showing an approximately normal distribution. 
Importantly, this indicates that reverse-direction loss is not universally higher than forward-direction loss. In fact, for over half of the texts, reverse prediction of the next tokens is comparatively easier.

As indicated in Figure~\ref{figure:diff_domain}, compared to web-scraped corpora such as Wikipedia and Common Crawl, the distributions of loss differences from Book and ArXiv are generally less skewed towards easier forward-modeling. Furthermore, a larger proportion of texts in Book and ArXiv are easier to predict in the reverse direction compared to the forward direction.
Considering that texts from books and academic papers are typically of higher quality than web-scraped texts, we speculate that texts, where reverse prediction is more effective, are generally more coherent, naturally flowing. Table~\ref{tab:main_cases} summarizes the randomly selected examples from the reverse easier and forward easier texts. The reverse easier texts display a coherent structure and smooth flow, making them easy for readers to follow. In contrast, the forward easier texts are relatively low-quality, less coherent, and often repetitive. This conjecture is also reflected in domains related to code, StackExchange, and Github. From the perspective of natural language, code often features monotonous syntax and repetitive vocabulary.

From the perspective of human forward thinking and its reflection in written texts, the forward-direction prediction task, which involves predicting the future from the present, is inherently more challenging. Conversely, the reverse-direction token prediction operates from known outcomes back to their origins, potentially simplifying the task. 

\subsection*{RQ3: \textit{What features make texts easier to process in the reverse direction?}}
\input{figures/dynamic_assumption_a}
\input{figures/step_loss}
To further validate our hypothesis, we conduct a detailed analysis of step-by-step loss changes during token decoding. While the aggregated view in RQ2 is informative, it hides the underlying step-by-step dynamics of the loss. Given $m$ text sequences in the SlimPajama dataset, we can obtain the two step-by-step loss sequences $\{\frac{\sum_{j=1}^m F_1^j}{m}, \frac{\sum_{j=1}^m F_2^j}{m}, \cdots, \frac{\sum_{j=1}^m F_N^j}{m}\}$ for the forward modeling and $\{\frac{\sum_{j=1}^m R_1^j}{m}, \frac{\sum_{j=1}^m R_2^j}{m}, \cdots, \frac{\sum_{j=1}^m R_N^j}{m}\}$ for the reverse modeling. We exclude the first and last tokens with step loss = $0$ to avoid sharp changes at the start and end. To account for different text lengths, we normalize the steps of all texts to the interval $(0, 1)$.

Given our findings that LLMs can effectively learn both forward and reverse modeling when trained from scratch, we initially hypothesize a straightforward relationship between the step-by-step loss of two directions, which is shown as assumption (a) in Figure~\ref{figure:assum_a}. The assumption (a) is that forward and reverse modeling would exhibit similar loss patterns throughout the sequence, explaining the near-zero mean difference in average losses in RQ2. However, our empirical results, presented in Figure~\ref{figure:step_loss}, reveal a more nuanced dynamic. The results instead support assumption (b) in Figure~\ref{figure:assum_a}: reverse prediction becomes progressively more accurate as contextual information accumulates, while forward prediction maintains consistent difficulty levels across the sequence. These trajectories intersect at a critical intersection point, before which reverse prediction shows higher loss values, and after which it demonstrates lower loss values compared to forward prediction. Note that this pattern emerges consistently across all the texts. It is a statistical characteristic of all the texts in our datasets and is independent of text quality, representing a fundamental difference of LLM’s forward- and reverse-modeling behaviors. 

To further understand this dyanmic, we analyze extreme cases (those in the top and bottom $10\%$ of average loss differences) to identify the features that drive these divergent patterns and to examine how these dynamics change in extreme cases. A straightforward hypothesis (assumption (c) in Figure~\ref{figure:assum_b}) would suggest that extreme cases simply shift the reverse loss curve vertically while maintaining its shape, with top-$10\%$ cases shifting upward and bottom-$10\%$ cases shifting downward. Under this hypothesis, the intersection point between forward and reverse loss curves would show small distance changes. However, our findings in Figure~\ref{figure:step_loss_split} contradict this hypothesis and instead support assumption (d) in Figure~\ref{figure:assum_b}: extreme cases primarily result in large horizontal shifts of the reverse loss dynamic, while the forward loss dynamic remains stable (simple vertical shift). In cases where forward loss substantially exceeds reverse loss (top-$10\%$ cases), we observe that reverse loss decreases rapidly, with the intersection point occurring very early in the sequence. Conversely, in cases where reverse loss is larger (bottom-$10\%$ cases), the intersection point is delayed until near the sequence end, with reverse loss consistently exceeding forward loss throughout most of the process.

\input{figures/dynamic_assumption_b}
\input{figures/step_loss_split}

\input{tables/llama_mmlu}

The results shows that while the average loss difference is an aggregated metric, it effectively indicates different patterns in step-by-step loss dynamics. With our case studies in Table~\ref{tab:main_cases}, we find the obvious text quality differences between the reverse-favoring cases and forward-favoring cases. This finding suggests that text quality is the key feature influencing the loss dynamics and the positions of intersection points. Our analysis also reveals a key insight: texts exhibiting early intersection points in their loss dynamics typically have higher loss differences and correspond to higher quality content. This relationship enables us to use the loss difference as a quality score for text quality assessment.


\subsection*{Application: \textit{Texts favoring reverse modeling can improve original LLMs.}}

\noindent As analyzed in RQ3, coherent and logical texts tend to have lower reverse losses compared to forward losses. Thus, given a training sample and a LLM model pre-trained from scratch with both forward and reverse training, we can define a simple quality score $\mathcal{S}$ using the loss difference \uline{$\mathcal{S} = \text{Avg. Forward Loss}$ - $\text{Avg. Reverse Loss}$}, computed as $\mathcal{S} = \frac{1}{N}(\sum_{i=1}^N F_i - \sum_{i=1}^N N_i)$.
According to our prior analysis, A higher $\mathcal{S}$ indicates that the text, which supports reverse modeling better, signifies a high-quality sample.

To further verify this assumption, we conduct continued pre-training on the publicly released Llama2-7B. Using the SlimPajama-6B~\cite{soboleva2023slimpajama} as training data, we select 1B tokens with the lowest and highest $\mathcal{S}$ scores, respectively. The model's performance is evaluated on MMLU~\cite{hendrycks2020measuring}. We also compare this with the following data selection strategies: (1) \textbf{Random 1B}: randomly sample 1B tokens, (2) \textbf{Perplexity Lowest Ranked 1B}: select the 1B tokens with the lowest perplexity by Llama2-7B.

The results from Table~\ref{table:llama_mmlu} show that the quality of training data significantly affects the performance of LLMs.
Our high-quality data selection strategy ($\mathcal{S}$ Highest Ranked) outperforms other baselines, achieving the highest accuracy across various tasks on MMLU.
Since the overall text quality of the SlimPajama 6B dataset is inferior to the text quality used in the pretraining of Llama2-7B,
using the full 6B dataset does not improve over the original Llama2-7B.
This suggests that the presence of low-quality data in unfiltered training sets degrades performance, as evidenced by the significant performance decline with low-quality selection strategy ($\mathcal{S}$ Lowest Ranked).
This experiment supports the hypothesis that texts more effectively modeled by reversing are of higher quality and more beneficial for LLMs in acquiring world knowledge.

%% file: figures/multilingual_loss_en.tex
\begin{figure}[!t]
\centering
\includegraphics[width=0.5\textwidth]{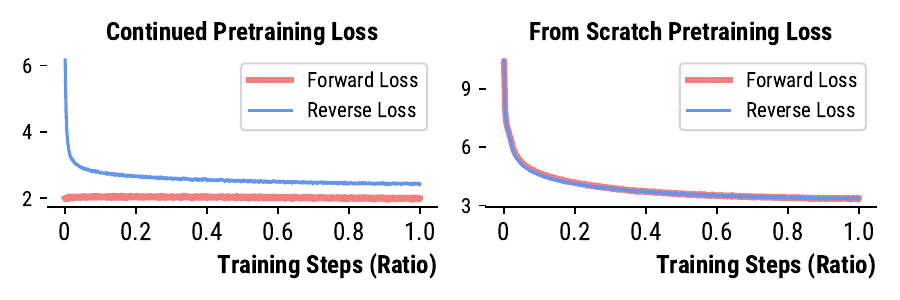}
\vspace{-2em}
\caption{
Pre-training loss for both continued setting and from-scratch settings in English.
}
\vspace{-1.5em}
\label{figure:multilingual_loss_en}
\end{figure}

%% file: tables/main_cases.tex
\begin{table*}[!t]
\centering
\vspace{-1em}
\fontsize{9}{10}\selectfont
\begin{tabularx}{\textwidth}{XX}
\toprule
\textbf{Text Favoring Reverse (Low Reverse Loss)} & \textbf{Text Favoring Forward (Low Forward Loss)} \\
\midrule
Whether you like it or not, your garden is an open park for all of nature’s creatures. ... Let’s take a few minutes to learn all about ladybugs in your garden.
Are Ladybugs Good for your Garden? ... Now that you know all about ladybugs and their role in controlling the aphid population, you may be interested in attracting ladybugs to your garden. ... & Ubuntu Manpage: \newline phm2helix - calculate projections through a time varying phantom object. ...  phm2helix - calculate projections through a time varying phantom object. ... phm2pj calculates projections through a time varying phantom object. ... \\
\bottomrule
\end{tabularx}
\vspace{-2mm}
\caption{
We sample one text favoring reverse and one favoring forward, using ``...'' to omit sentences while preserving the main structure. Texts favoring reverse are often structured with clear logic flows, but texts favoring forward rely heavily on formatting to convey their sequential flow. 
More multilingual cases are shown in the Appendix~\ref{app:multilingual}.
}
\label{tab:main_cases}
\end{table*}

%% file: figures/slim_diff_loss_domain_distribution.tex
\begin{figure}[t!]
\centering
\includegraphics[width=0.48\textwidth]{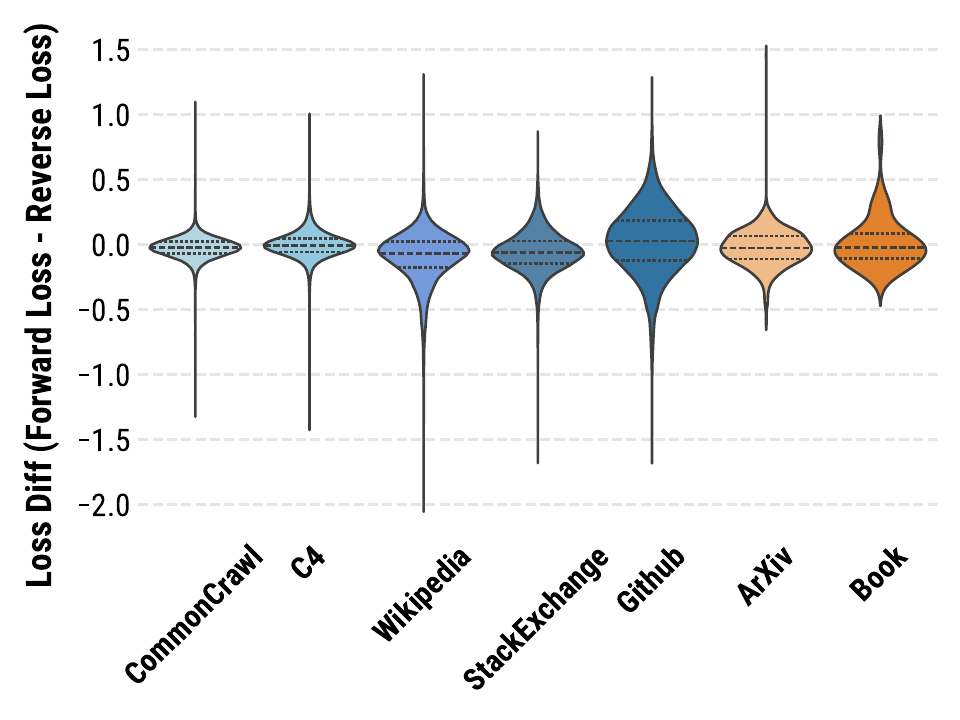}
\vspace{-2.6em}
\caption{
Loss difference distribution across domains.
}
\vspace{-1.5em}
\label{figure:diff_domain}
\end{figure}

%% file: figures/dynamic_assumption_a.tex
\begin{figure}[!t]
\centering
\includegraphics[width=0.5\textwidth]{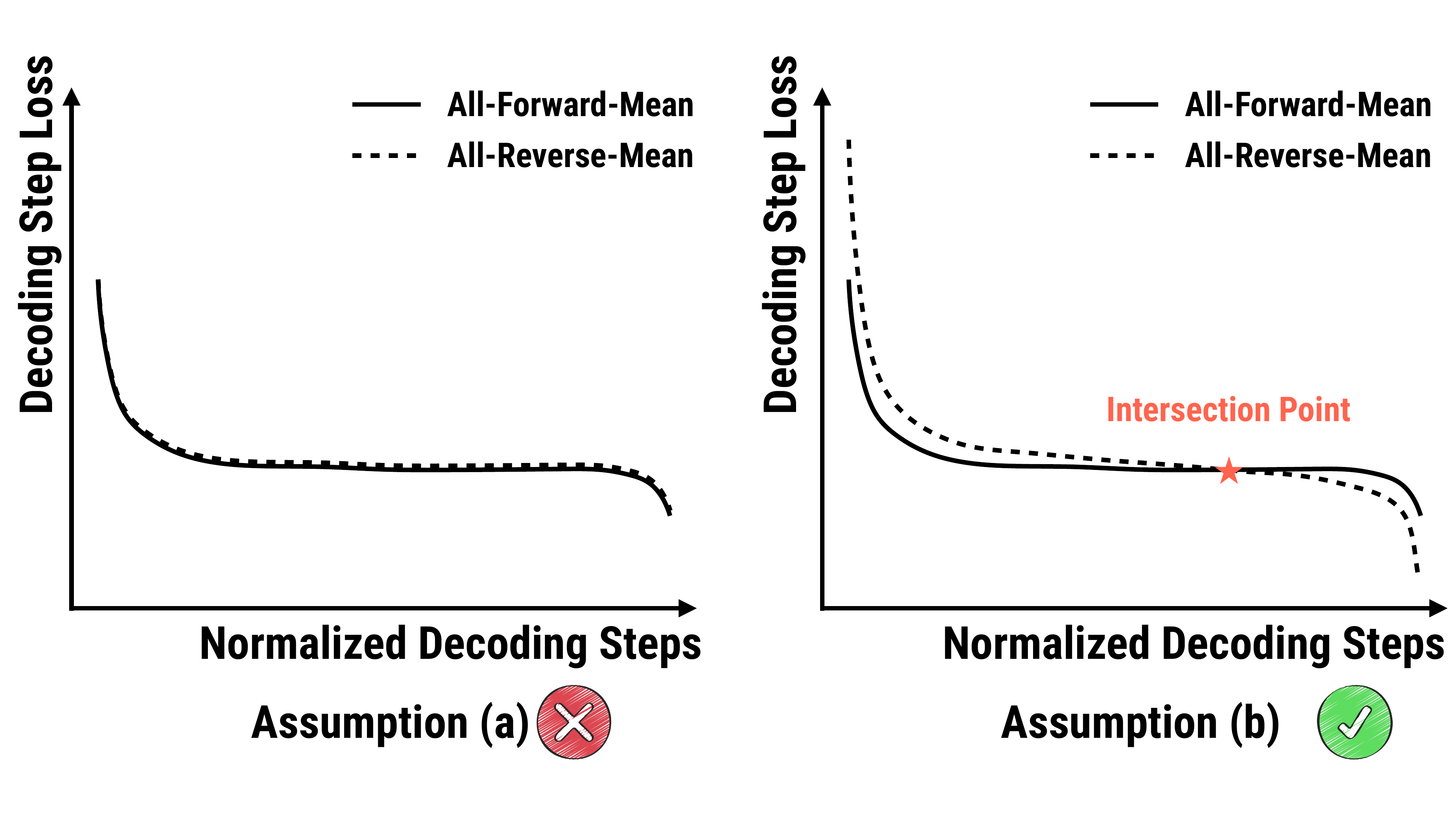}
\vspace{-2em}
\caption{
Assumptions on the step-by-step loss dynamics of full text data during decoding.
}
\vspace{-1em}
\label{figure:assum_a}
\end{figure}

%% file: figures/step_loss.tex
\begin{figure}[t!]
\centering
\includegraphics[width=0.5\textwidth]{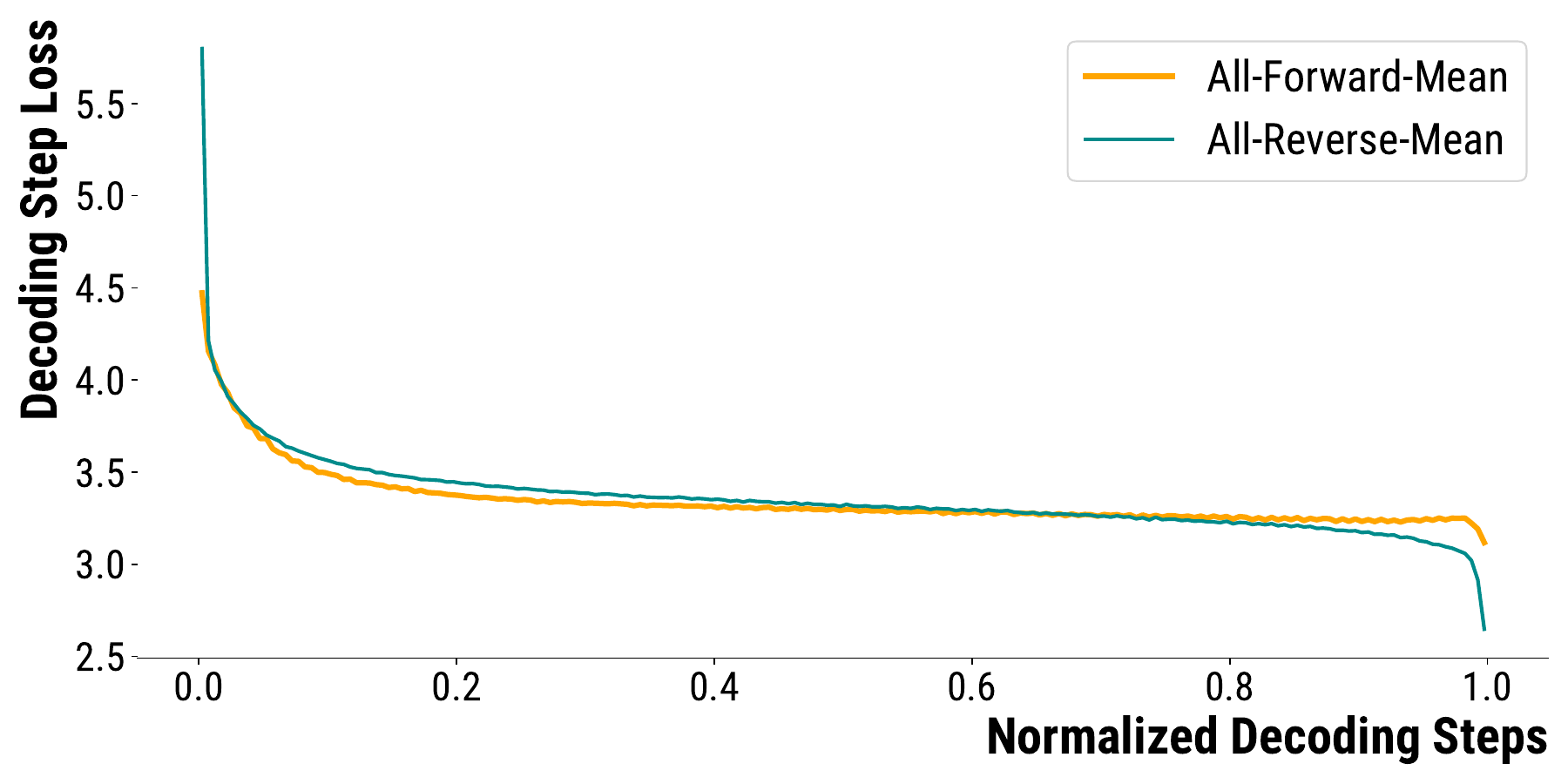}
\caption{
Empirical step-by-step loss dynamics of full text data during decoding.
}
\vspace{-1.5em}
\label{figure:step_loss}
\end{figure}

%% file: figures/dynamic_assumption_b.tex
\begin{figure}[!t]
\centering
\includegraphics[width=0.5\textwidth]{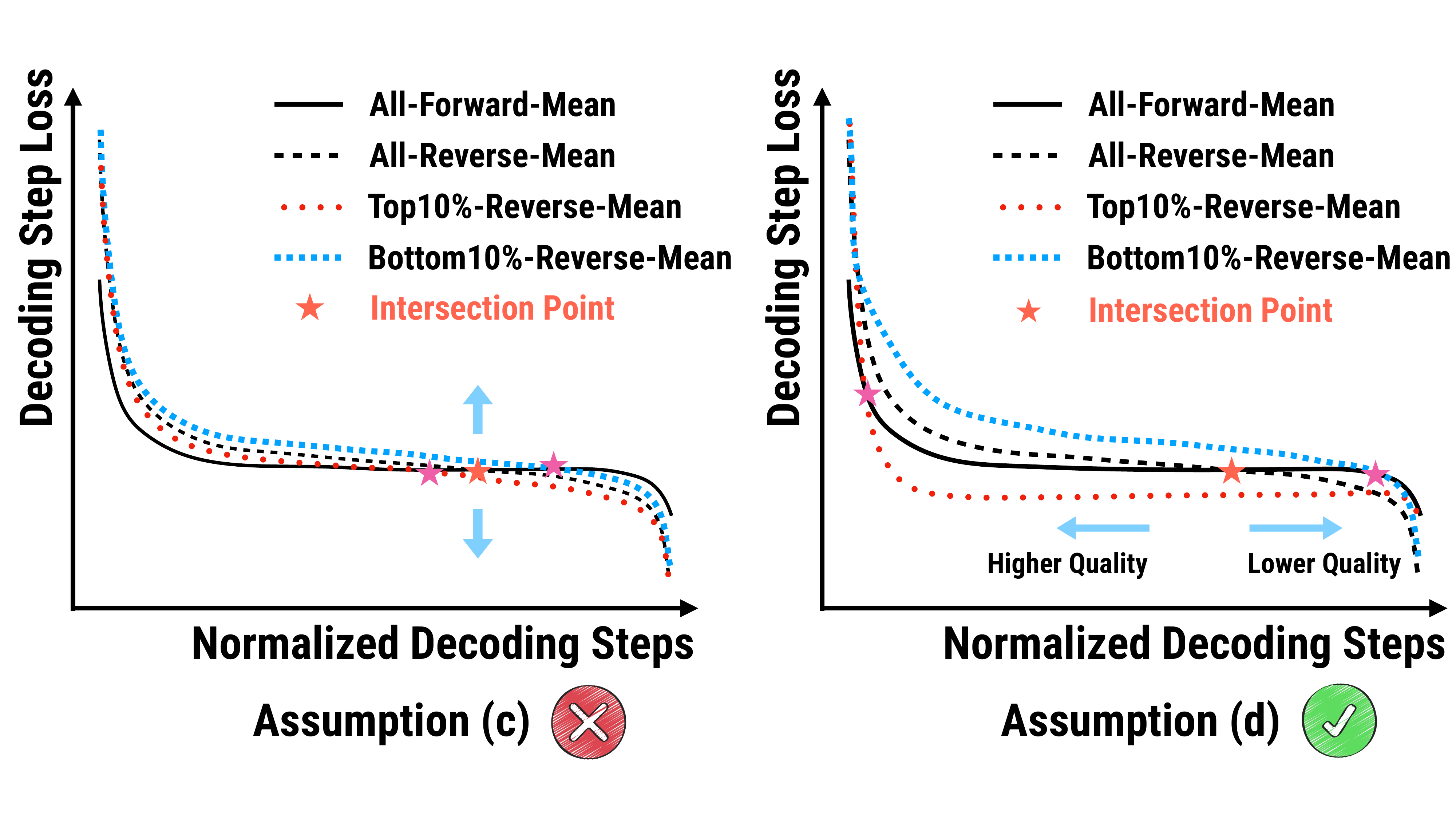}
\vspace{-2em}
\caption{
Assumptions on the step-by-step loss dynamics of selected texts with the Top-$10\%$ and Bottom-$10\%$ loss differences during decoding.
}
\vspace{-1em}
\label{figure:assum_b}
\end{figure}

%% file: figures/step_loss_split.tex
\begin{figure}[t!]
\centering
\includegraphics[width=0.5\textwidth]{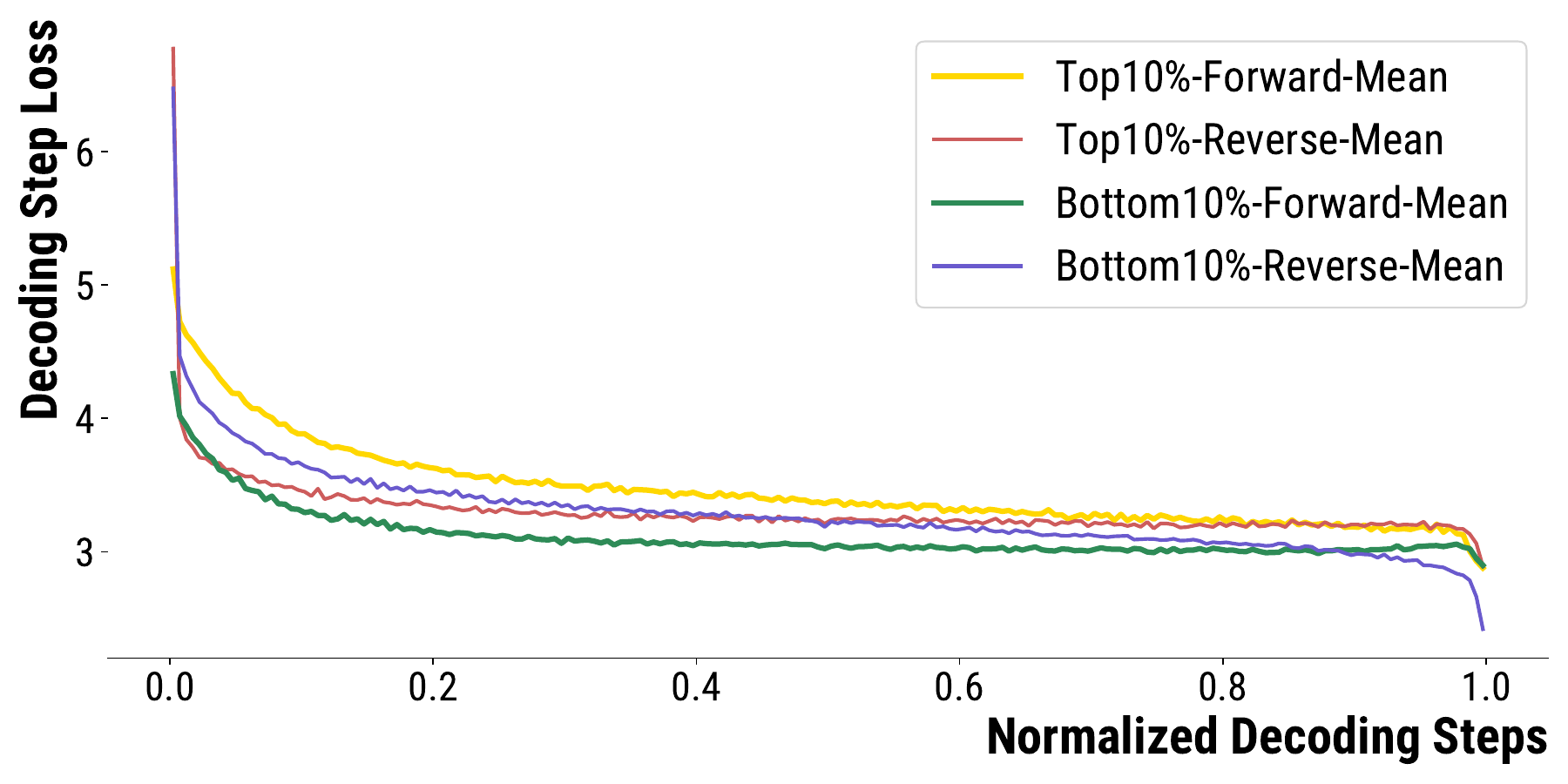}
\caption{
Empirical step-by-step loss dynamics of selected texts with the Top-$10\%$ and Bottom-$10\%$ loss differences during decoding.
}
\vspace{-1.5em}
\label{figure:step_loss_split}
\end{figure}

%% file: tables/llama_mmlu.tex
\begin{table*}[!htbp]
\centering
\fontsize{9}{10}\selectfont
\begin{tabular}{l|cccc|c}
\toprule
\textbf{Model \& Strategy} & \textbf{Stem} & \textbf{Humanities} & \textbf{Social Science} & \textbf{Other} & \textbf{Average} \\
                            \cmidrule{1-6}
Original Llama2-7B    & 35.84       & 50.60       &    50.46   &     48.10      &   45.29         \\       
CT w/ All SlimPajama-6B          &   36.15     & 46.74       &   49.03    & 46.63          &   43.85      \\ 
CT w/ Random 1B          &   35.73     & 46.16       &   48.40    & 47.08          &   43.57      \\ 
CT w/ PPL Lowest Ranked 1B          &   36.24     & 45.79       &   47.57    & 45.53          &   43.09      \\ 
CT w/ $\mathcal{S}$ Lowest Ranked 1B          &   34.04     & 45.94       &   45.66   & 42.93         &   41.38      \\ 
CT w/ $\mathcal{S}$ Highest Ranked 1B          &   \textbf{37.15}     & \textbf{50.93}       &   \textbf{50.63}   & \textbf{49.82}         &   \textbf{46.24}      \\ 
\bottomrule   
\end{tabular}
\vspace{-2mm}
\caption{
Results (Accuracy$\%$) on the MMLU benchmark among different data selection strategies on LLaMA2-7b continued pre-training (CT). $\mathcal{S}$ is our proposed quality score simply computed by $\text{Forward Loss}$ - $\text{Reverse Loss}$. More results with different backbones across various benchmarks are shown in Appendix~\ref{app:more}.}
\vspace{-0.5em}
\label{table:llama_mmlu}
\end{table*}

%% file: sections/5_con.tex
\section{Conclusions}
In conclusion, our results demonstrate that LLMs can learn from both forward and reverse-modeling texts with comparable proficiency when trained from scratch.
This study also highlights the potential benefits of incorporating training data that favors reverse modeling. 
Our findings underscore the importance of exploring diverse reverse modeling frameworks to enhance the capabilities of LLMs.

%% file: sections/6_limitation.tex
\section*{Limitations}
While our study demonstrates promising results in training LLMs with reverse modeling, several limitations should be acknowledged to provide a comprehensive understanding of the findings and guide future research.

Firstly, the simulation of reverse modeling by simply reversing token sequences may not fully capture the complexity and nuances of true reverse thinking processes. This approach reduces reverse modeling to a syntactic level, potentially overlooking deeper semantic and contextual factors intrinsic to human reverse modeling.

Secondly, the evaluation metrics used in our study, such as performance on downstream benchmarks like MMLU, may not fully encompass the benefits or limitations of reverse modeling. These metrics primarily measure specific aspects of language understanding and reasoning, potentially overlooking other critical dimensions influenced by reverse modeling, such as creativity or problem-solving skills.

Lastly, our research does not address the potential computational and resource challenges associated with training LLMs on reverse texts. The increased complexity and processing demands could pose significant barriers to practical applications, particularly in resource-constrained environments.

In conclusion, while our findings offer valuable insights into the potential of reverse modeling in LLMs, addressing these limitations is crucial for advancing this line of research. Future studies could aim to develop more sophisticated methods for simulating reverse modeling, explore diverse and naturally occurring datasets, and consider a broader range of evaluation metrics to fully understand and harness the benefits of reverse modeling in LLMs.

%% file: sections/7_appendix.tex
\clearpage
\renewcommand\thesection{\Alph{section}}
\setcounter{section}{0}

\section{Implementation Details}
\label{app:implement}
In our initial experiments, we explore the effects of a higher learning rate ($2e^{-4}$) and extend the training epochs ($2$ epochs) in a continued training setting. This exploration is based on the assumption that reverse modeling might need more epochs or a higher learning rate than forward modeling to overcome the pre-trained directional bias. While this increases the convergence speed, the final loss is nearly the same as when using a lower learning rate or a single training epoch.

Thus, to ensure consistency in our comparisons, we fix the learning rate as $5e^{-5}$ and set the batch size to $48$ (with each batch consisting of $48$ paragraphs). Following established practices in LLM training~\cite{touvron2023llama,chowdhery2023palm}, we train for one epoch in all experiments. The number of training steps depends on the size of the data set and the batch size. For instance, training on a dataset with $1$ billion tokens requires approximately $2,400$ steps with our hyperparameter settings.

We use Llama2-7B~\cite{touvron2023llama} as the LLM backbone for Research Questions 1–3 (Section~\ref{section:exp}). All experiments are conducted using $8$ NVIDIA A100-SXM-80GB GPUs, and the application experiments in Section~\ref{section:exp} also use the Llama2-7B model.

\input{figures/multilingual_loss}
\section{Multilingual Experimental Results}
\label{app:multilingual}
We show the pre-training losses for both continued and from-scratch training across additional languages, including German, Korean and Arabic, in Figure~\ref{figure:multilingual_loss}. Note that Arabic texts tokenized by Llama2 tokenizer have the same orientation as English, with tokens from the first logical sentence of a paragraph positioned on the left rather than the right. Consistent with our findings in RQ1, Section~\ref{section:exp}, the forward loss during continued pre-training remains lower than the reverse loss. However, in the from-scratch setting, the loss curves for both directions converge similarly. These results further confirm that LLMs can effectively learn to handle both forward and reverse inputs with comparable proficiency when trained from scratch, regardless of languages.

We randomly sample additional multilingual cases (German, Korean, and Arabic), as shown in Tables 4-7. Across all four languages, we observe a consistent pattern: Texts favoring reverse modeling tend to exhibit clear logical structures, while those favoring forward modeling rely more on repetitive formatting to emphasize their sequential flow.

\section{More Results with Different LLM Backbones on Different Benchmarks}
\input{tables/more_benchmarks}
\label{app:more}
Besides the MMLU~\cite{hendrycks2020measuring} benchmark, we also evaluate our proposed data selection application on three benchmarks, \ie, AGIEval~\cite{zhong2023agieval}, BBH~\cite{suzgun2023challenging} and BoolQ~\cite{clark2019boolq}, using different LLM backbones including Llama2-7b~\cite{touvron2023llama}, Mistral-7b~\cite{jiang2023mistral} and Llama3-8b~\cite{dubey2024llama}.

The experimental results are shown in Table~\ref{tab:more_benchmarks}. Our high-quality data selection strategy (Highest Ranked) consistently outperforms other approaches across various benchmarks, regardless of the LLM backbone used. These results support our hypothesis that texts better modeled by reverse prediction yield higher quality data, which in turn enhances the LLMs' ability to acquire world knowledge. Notably, the ranked score, $\mathcal{S} = \text{Forward Loss}$ - $\text{Reverse Loss}$, is computed and fixed using the Llama2-7b model throughout the experiments. This highlights the strong generalization capability of our method, as the high-quality data selected by one LLM can be effectively transferred to another.

\input{figures/gpt4-prompt}
\section{Qualitative Analysis}
Many open-source LLMs use data classifiers for selection and cleaning, but the specifics of their training processes are often proprietary and not fully detailed in technical reports~\cite{touvron2023llama,jiang2023mistral,dubey2024llama}. Thus, we expand our experiments using the GPT-4 API~\footnote{\url{https://platform.openai.com/docs/api-reference/}} to directly evaluate the quality of texts from the Lowest Ranked 1B and Highest Ranked 1B texts, in line with previous studies~\cite{clark2021all,cornelius2024bust}. We randomly select $1,000$ samples from each dataset and apply a predefined prompt (shown in Figure~\ref{figure:gpt4-prompt}) to assess each sample. The GPT-4 API assigns a quality score ranging from $1$ to $10$, based on criteria for high-quality text defined in~\cite{clark2021all}, along with an additional criterion for ``quality for LLM training''.

The results show that the Highest Ranked 1B dataset achieves an average score of $6.7$, while the Lowest Ranked 1B and Random 1B datasets scored $4.9$ and $6.15$, respectively. These findings further suggest that texts favoring reverse modeling are of higher quality and more suitable for LLM training.

\section{Smaller LLM as Backbone Model}
\input{figures/tinyllama_fig}
We also conduct experiments on the TinyLlama-1.1B model~\cite{zhang2024tinyllama} under the same protocols used in Section 4~\ref{section:exp}. Figure~\ref{figure:tinyllama_en} shows that the loss trend of TinyLlama-1.1B is similar to that of the larger Llama2-7B model. It indicates that even with smaller models, both forward and reverse modeling can be effectively handled in the from-scratch training setting. Next, we recalculate the quality score $\mathcal{S} = \text{Forward Loss}$ - $\text{Reverse Loss}$ for each case of the SlimPajama~\cite{soboleva2023slimpajama} dataset and select the lowest-ranked 1B and highest-ranked 1B data based on the recalculated scores. Notably, the overlap ratios of the selected cases are $91.27\%$ and $94.58\%$ when compared to Llama2-7B. This demonstrates that our method is model-agnostic, enabling smaller models to efficiently identify high-quality data for training larger models in practice.

\clearpage
\input{tables/more_cases_en}
\clearpage
\input{tables/more_cases_de}
\clearpage
\includepdf[pages=1]{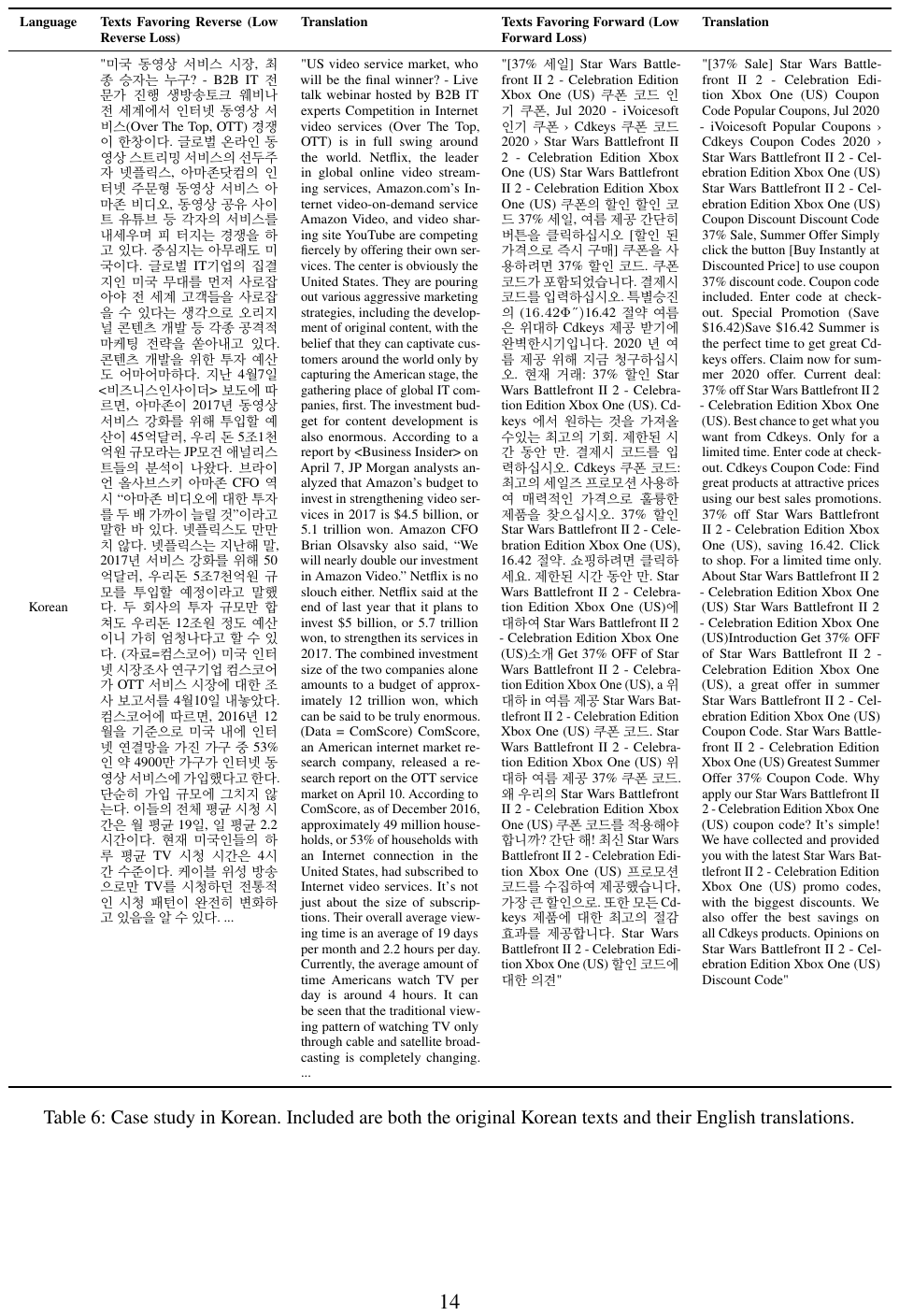}
\clearpage
\includepdf[pages=1]{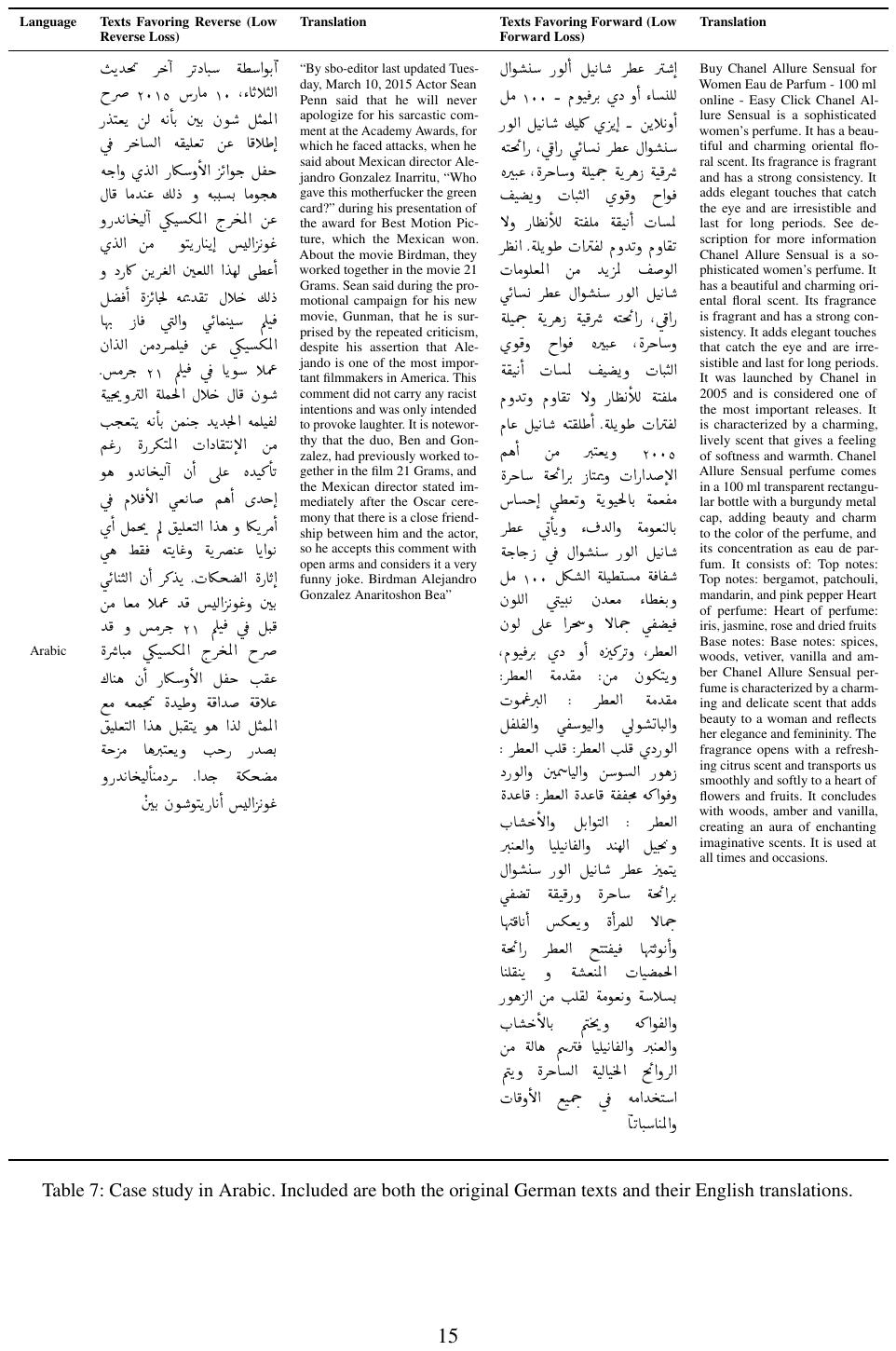}

%% file: figures/multilingual_loss.tex
\begin{figure*}[h]
    \centering
    \includegraphics[width=0.98\textwidth]{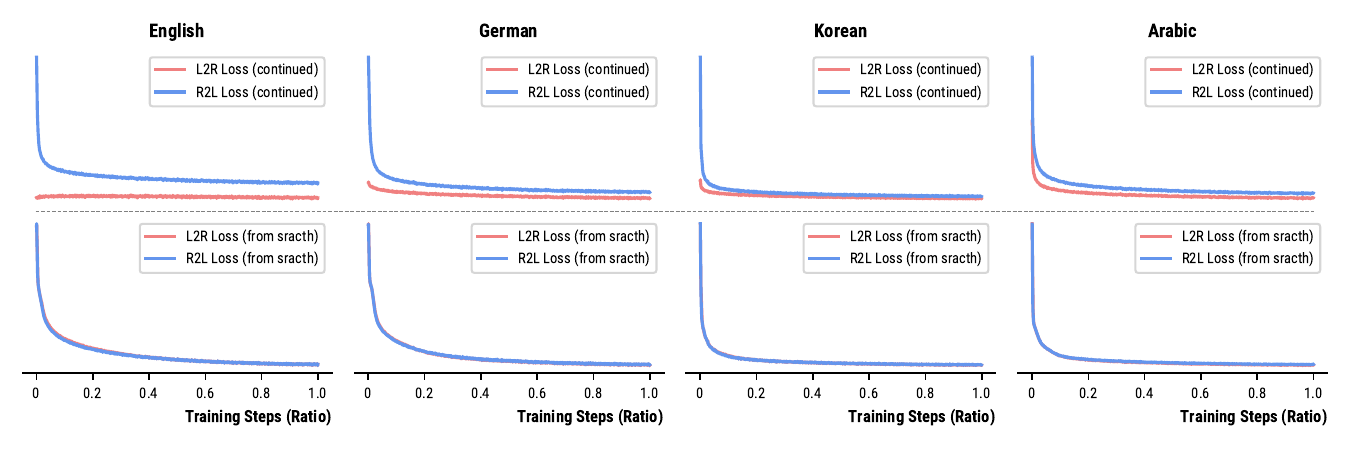}
    \caption{
    Pre-training losses for both continued and from-scratch training settings in four additional languages. The patterns are consistent with the results observed in English.
    }
    \label{figure:multilingual_loss}
\end{figure*}

%% file: tables/more_benchmarks.tex
\begin{table}[t!]
\setlength\tabcolsep{4pt}
\small
\begin{tabular}{lcccc}
\toprule
\textbf{Model \& Strategy} & \textbf{MMLU}  & \textbf{AGIEval} & \textbf{BBH}   & \textbf{BoolQ} \\\midrule
\multicolumn{5}{l}{\textbf{\textit{Llama2-7B}}} \\
\cmidrule{0-0}
Random 1B  & 43.57          & 26.53            & 42.33          & 74.86          \\
Lowest Ranked 1B  & 41.38   & 25.56            & 38.26          & 73.96          \\
Highest Ranked 1B & \textbf{46.24} & \textbf{27.07}   & \textbf{43.79} & \textbf{75.44} \\\midrule
\multicolumn{5}{l}{\textbf{\textit{Mistral-7B}}} \\
\cmidrule{0-0}
Random 1B   & 35.45          & 40.97            & 43.45          & 77.34          \\
Lowest Ranked 1B  & 34.99    & 38.85            & 42.17          & 75.29          \\
Highest Ranked 1B & \textbf{36.66} & \textbf{42.86}   & \textbf{44.98} & \textbf{78.56} \\\midrule
\multicolumn{5}{l}{\textbf{\textit{Llama3-8B}}} \\
\cmidrule{0-0}
Random 1B &  58.94      & -       & -       & -        \\
Lowest Ranked 1B  & 58.54   & - & -  & -      \\
Highest Ranked 1B & \textbf{59.49} & - & -& -    \\\bottomrule
\end{tabular}
\caption{Experimental results using three different LLM backbones on the MMLU, AGIEval, BBH, and BoolQ benchmarks. However, we exclude Llama3-8B's results on AGIEval, BBH, and BoolQ, as the evaluation sets for these benchmarks are found to overlap significantly with its training data (contaminated rates: $98\%$, $95\%$, and $96\%$, respectively)~\cite{dubey2024llama}.}
\label{tab:more_benchmarks}
\end{table}

%% file: figures/gpt4-prompt.tex
\begin{figure*}[th]
    \centering
    \includegraphics[width=0.80\textwidth]{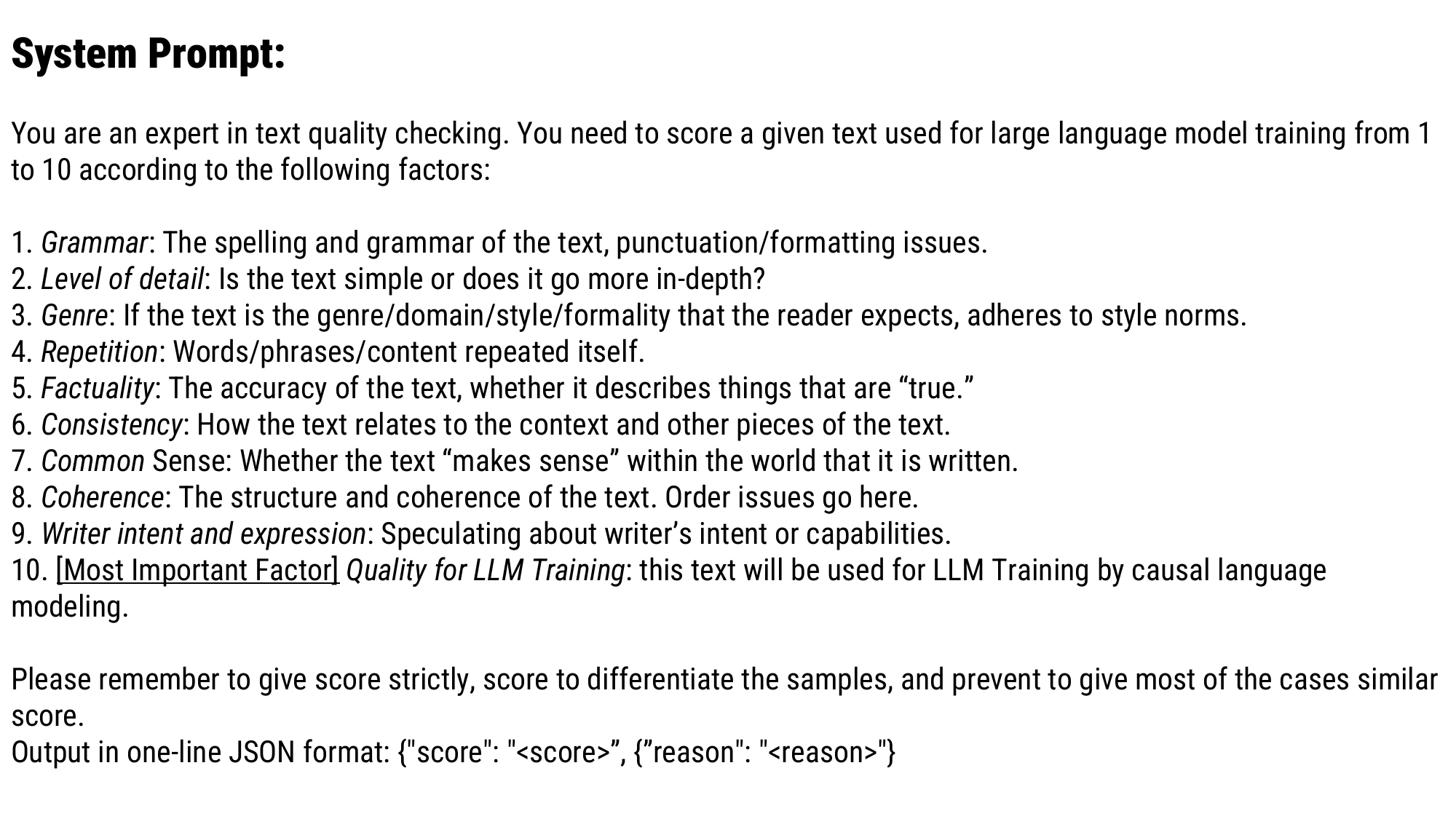}
    \caption{
    The prompt used for text qualitative evaluation using GPT-4 API. The $1$-$9$ factors follows the designed evaluation labels in~\cite{clark2021all}, and we add an extra ``Quality for LLM Training'' into the evaluation factors.
    }
    \label{figure:gpt4-prompt}
\end{figure*}

%% file: figures/tinyllama_fig.tex
\begin{figure}[th]
\centering
\includegraphics[width=0.50\textwidth]{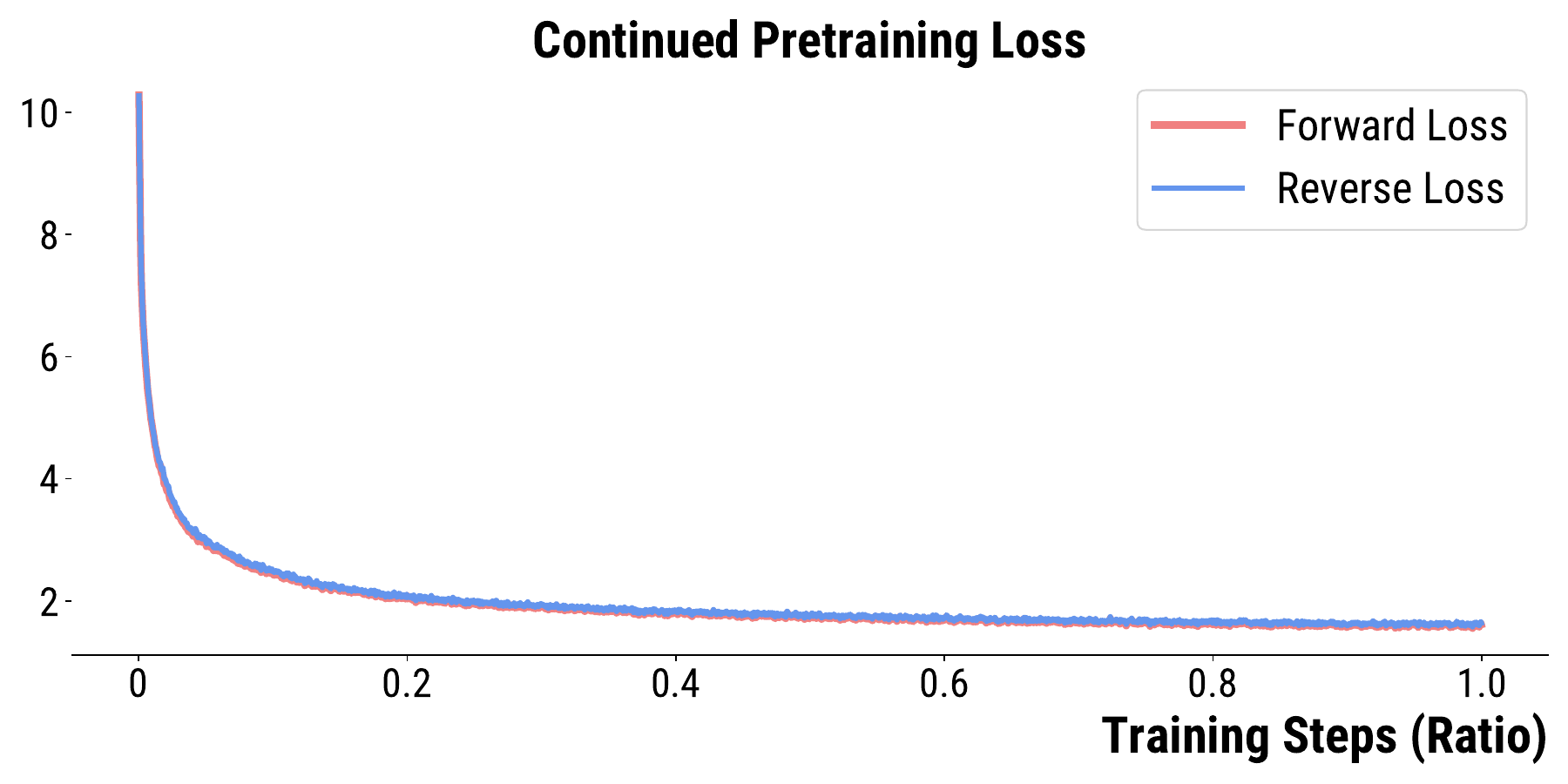}
\caption{
Pre-training loss in the from-scratch setting using English data on TinyLlama-1.1B. The forward and reverse losses are nearly identical, aligning with findings from larger LLM models.}
\label{figure:tinyllama_en}
\end{figure}


%% file: tables/more_cases_en.tex
\begin{table*}[!t]
\scriptsize
\begin{tabularx}{\textwidth}{cXX}
\toprule
\textbf{Language} & \textbf{Texts Favoring Reverse (Low Reverse Loss)} & \textbf{Texts Favoring Forward (Low Forward Loss)} \\
\midrule
\multirow{4}{*}[-42em]{English} & Have you ever wondered why cultures in the hottest locations on earth eat hot and spicy foods? Why is it that people in Central and South America, India, Africa, Southeast Asia and the Caribbean eat foods flavored with hot chile peppers and spices that make you sweat? There is a reason, and it's actually pretty smart when you think about it — spicy foods make you sweat, which in turn helps you cool down faster. It’s as simple as that!
Though you may be inclined to cool down with a tall glass of iced tea, ice cream or watermelon on a sweltering summer's day, the effect isn't lasting. After a while you're back to where you started — hot and bothered. That's because your internal temperature is cooled too rapidly, and your body ends up compensating by raising your temperature. As a result, you feel hotter.
Eating spicy foods works differently — it raises your internal temperature to match the temperature outside. Your blood circulation increases, you start sweating and once your moisture has evaporated, you've cooled off. Scientists call the phenomenon ``gustatory facial sweating,'' because indeed you usually start sweating in the face first.
Even though eating spicy foods on a hot day isn't the most pleasant for many people, it may be worth doing because after sweating it out you do actually cool down. What do you think: is it worth it? & GET \$2000 CASHBACK!!!! Nissan Qashqai combines stunning looks, efficient aerodynamics, and advanced technology to help you enjoy enlightened driving at its best. And thanks to Nissan Intelligent Mobility, you’ll feel more confident and connected than ever. Loaded with the state of the art features including: -5 Star ANCAP Safety Rating -Forward-Collision Warning -Flat-Bottom Steering Wheel -Black Leather-accented Seat and Steering Wheel Trim -Individually Heated Front Seats -Dual Zone Climate Control -7?809D Touch Screen Display -Satellite Navigation -Digital (DAB ) Radio -Intelligent Around-View Cameras -Blind Spot Alert -Rear Cross-Traffic Alert -Rear Privacy Glass -Fog Lights -Roof Rails -18?809D alloy wheels -LED Daytime Running Lights and Taillights -ISOFIX Child Restraint anchorage -Vehicle Dynamic Control -Cruise control with digital speedometer -Bluetooth hands free system with audio streaming -6 speaker sound system -AUX/iPod connectivity -Power windows -Power mirrors and much more! Located on Road in , close to public transport and freeways, and only a 25 minute drive from the CBD, we have been selling and servicing Nissan vehicles across Melbourne for over 25 years. \\
\cmidrule{2-3}
 & The kinetic equations for clean superconductors ($I \gg \xi$) are derived. Expanding the equations for the time dependent Green functions in the quasiclassical parameter, the new contributions are found which contain the derivatives of the distribution functions with respect to the quasiparticle momentum. The transition from the ultra-clean case (no relaxation) to a relaxation-dominated behavior, for which the kinetic equations coincide with the usual quasiclassical approximation, occurs for the relaxation time of the order of $\hbar EF/\Delta 2$. The kinetic equations can be used for various dynamic processes in superconductors including the flux-flow Hall effect. The derived equations, after necessary modifications for the p-wave pairing, are especially suitable for nonstationary problems in the theory of superfluidity of 3He. & Our Cartridges for Lexmark X2250 are great value with super fast delivery!
Cartridges for Lexmark X2250 are among our thousands online products. With our huge range and simple website, it is easy to find all the cartridges you need for any other printers you may have. Together with our most competitive prices, we are sure to be your one-stop online store!
Cartridges for Lexmark X2250 are covered by a 60 days warranty. If the product you received is faulty, please contact us to organise a replacement or refund. Please refer to our Warranty Return.
When will my Cartridges for Lexmark X2250 be delivered?
In most cases you will receive your Cartridges for Lexmark X2250 the next working day, or within 3 days if outside the next day express post network. It might takes up to 6 days for some remote areas. \\
\cmidrule{2-3}
 & How is this club different from standard Toastmasters Clubs?
As an Advanced Toastmasters Club, Professional Speakers Frankfurt is only open to experienced members who have completed the Toastmasters Competent Communicator level or are advanced speakers with proven experience outside the Toastmasters world. Therefore, we can focus on more advanced issues.
Rather than having the majority of speeches from the Competent Communicator manual we focus on advanced projects or practice speeches outside of Toastmasters manuals. We prepare members for speech competitions or help someone with an upcoming professional or other important speech.
Instead of having only one evaluator, all attendees will get the chance to provide feedback. Depending on the objectives of the speaker, the group might be divided into task forces to keep an eye on particular aspects and debrief the speaker afterward. We also use video recording to provide in-depth analysis of a speaker's performance.
Sounds boring? In fact, it isn't. For a speech to become great, it has to go through multiple iterations. In our club, we give members the opportunity to repeat a speech and incorporate the feedback they've received.
We hold regular advanced workshops run by members or outside experts on specific speech-related topics.
We encourage our members to participate in Toastmasters speech contests and dedicate special time to prepare candidates.
We will develop a peer coaching system through which members continuously coach each other.
We will set up a Speakers Bureau, and members will be able to present and promote themselves and as speakers on the Club website and through the Club’s online channels. & Rent a Dumpster in Oswego Now!
Simply give us a call and we will answer any questions you may have about the Oswego dumpster rental procedure, allowed materials, size you may need, etc.
Our roll off containers can usually be delivered in Oswego the day after you place your order. Make sure to call us early to ensure timely delivery.
Whether or not you require a long-term or roll-off dumpster is dependent upon the type of job and service you need. Long-Term dumpster service is for ongoing demands that last more than simply a few days. This includes matters like day-to-day waste and recycling needs. Temporary service is precisely what the name suggests; a one time need for project-special waste removal.
Temporary roll off dumpsters are delivered on a truck and are rolled off where they'll be utilized. These are typically larger containers that may manage all the waste that comes with that specific job. Long-Term dumpsters are generally smaller containers because they're emptied on a regular basis and so don't need to hold as much at one time.
Should you request a permanent dumpster, some firms require at least a one-year service agreement for this dumpster. Rolloff dumpsters only require a rental fee for the time that you keep the dumpster on the job.
If you want to rent a dumpster in Oswego, you will find that costs vary significantly from state to state and city to city. One means to get genuine estimates for the service you need would be to telephone a local dumpster company and ask regarding their costs. You can also request a quote online on some sites. These sites may also contain full online service that is constantly open. On these sites, you can choose, schedule and pay for your service whenever it's convenient for you.
Variables which affect the price of the container contain landfill fees (higher in some areas than others) as well as the size of the container you opt for. You also need to consider transportation costs as well as the kind of debris you will be placing into your container. \\
\cmidrule{2-3}
& Complete replacement of factory floor automation systems program for the largest auto plant in North America. It was a body assembly plant, a paint plant and a final assembly plant. The size of the plant was being doubled to accommodate a new model. With less than three months to go, the launch was in jeopardy because the systems were not ready for installation. Failure to install would delay the new model six months. An unsuccessful installation would shut the plant down. & With SoundcloudToMp3 you can convert and download music in High Quality MP3 format.
Download tons of music from Soundcloud with our Soundcloud Downloader and listen to them from anywhere by storing them on your iPod, computer or phone using our ultra fast downloading service.
SoundCloud is audio distribution site, where users can record, upload and promote their sound tracks. SoundCloud allows you to listen as many tracks you can but it does not allow sound track download.
Enter the Soundcloud URL that you wish to convert \& Download.
Click "Convert it" to start the conversion process.
Click "Download Mp3" to download the file.
Once complete you will have final download link for converted sound.
Highly Secure and high speed.
Mp3 Converter supports a wide variety of modern browsers and devices.
\\
\bottomrule
\end{tabularx}
\caption{Case study in English. The texts favoring reverse are typically high-quality and well-suited for LLM training. In contrast, those favoring forward modeling often exhibit repetition and occasional lapses in logic and coherence, which can negatively impact LLM training.}
\label{tab:more_cases_en}
\end{table*}

%% file: tables/more_cases_de.tex
\begin{table*}[!t]
\scriptsize
\begin{tabularx}{\textwidth}{cXXXX}
\toprule
\textbf{Language} & \textbf{Texts Favoring Reverse (Low Reverse Loss)} & \textbf{Translation} & \textbf{Texts Favoring Forward (Low Forward Loss)} & \textbf{Translation} \\
\midrule
\multirow{2}{*}[-47em]{German} & 
In aller Munde, in aller Ohren – an Jonas Kaufmann kommt man derzeit nicht vorbei. Startenor, Herzensbrecher, ein echtes Münchner Kindl noch dazu, hat sich Kaufmann in die internationale erste Riege gesungen. „Seine Intensität und seine Eleganz, die Geschmeidigkeit seiner Stimme und seiner Körpersprache, kombiniert mit seiner Musikalität und seinem strahlenden Aussehen, machen ihn zum Inbegriff des Opernstars im 21. Jahrhundert“, schwärmte der Herausgeber der Opera News. Und so wird Jonas Kaufmann seit geraumer Zeit weltweit gefeiert – nicht nur an den größten Opernhäusern, sondern auch als Protagonist in Gustav Mahlers „Lied von der Erde“, als Interpret von Hugo Wolfs „Italienischem Liederbuch“ oder als leidenschaftlicher Tenor, wenn er in einer Hommage an die unsterbliche Musik Italiens ihren Evergreens eine besondere Magie verleiht. ...
& On everyone's lips, in everyone's ears – it's impossible to overlook Jonas Kaufmann at the moment. Star tenor, heartthrob, and a true Munich native, Kaufmann has sung his way into the international top ranks. 'His intensity and elegance, the smoothness of his voice and body language, combined with his musicality and his radiant appearance, make him the epitome of the 21st-century opera star,' enthused the editor of Opera News. And so, Jonas Kaufmann has been celebrated worldwide for quite some time – not only at the greatest opera houses, but also as the lead in Gustav Mahler's ``Das Lied von der Erde'', as an interpreter of Hugo Wolf's ``Italian Songbook'', or as a passionate tenor when he lends a special magic to Italian evergreens in a tribute to the immortal music of Italy. ...
& ...
Urlaubsangebote für Yaroslavl
Spielen Sie mit dem Gedanken, eine Reise nach Yaroslavl zu buchen? Ob Sie einen Romantikurlaub, eine Familienreise oder ein All-Inclusive-Paket planen, die Pauschalreisen nach Yaroslavl auf TripAdvisor machen die Reiseplanung einfach und erschwinglich.
Vergleichen Sie Hotel- und Flugpreise für Yaroslavl und finden Sie so auf TripAdvisor die perfekte Pauschalreise nach Yaroslavl. Reisende wie Sie haben 7.983 Bewertungen geschrieben und 10.284 authentische Fotos für Hotels in Yaroslavl gepostet. Buchen Sie Ihren Urlaub in Yaroslavl noch heute!
Familienfreundliche Hotels in Yaroslavl
“Gute Lage, ein Park und Kotorosl Ufer fußläufig gut erreichbar. Zimmer sind sauber und werden immer gut aufgeräumt. Ein sehr bequemes Bett, das man sehr selten findet. Auch einen sehr guten und ... 
& ... Holiday Offers for Yaroslavl
Are you thinking about booking a trip to Yaroslavl? Whether you are planning a romantic getaway, a family trip, or an all-inclusive package, the vacation packages to Yaroslavl on TripAdvisor make planning your trip easy and affordable.
Compare hotel and flight prices for Yaroslavl, and find the perfect package on TripAdvisor. Travelers like you have written 7,983 reviews and posted 10,284 authentic photos of hotels in Yaroslavl. Book your vacation to Yaroslavl today!
Family-Friendly Hotels in Yaroslavl
``Good location, with a park and the Kotorosl Riverbank within walking distance. The rooms are clean and always well-maintained. A very comfortable bed, which is hard to find. Also, a very good and...'' ...
\\
\cmidrule{2-5}
& ... Elisabeth von Luxemburg wurde 1422 13jährig mit dem 25 Jahre alten Thronanwärter Albrecht V. verheiratet (verlobt waren sie bereits seit ihrem 2. Lebensjahr). Nach den ersten zehn Jahren Ehe bekam sie ihr erstes von vier Kindern; fünf Jahre später wurde ihr Gemahl durch den Tod seines Vaters römisch-deutscher König sowie König von Ungarn, Kroatien und Böhmen. Elisabeth war im fünften Monat mit dem vierten Kind schwanger, als er 1439 während eines Feldzuges gegen die in Ungarn einfallenden Türken an der Ruhr verstarb. Entgegen dem politischen Drängen des Adels, den 15jährigen polnischen König Wladislaw III. zu heiraten – weil ein männlicher König gleich welchen Alters und Charakters für das Land im Krieg gegen die Türken „sicherer“ sei –, ergriff sie selbst die Regentschaft, um so bald als möglich ihren Sohn Ladislaus Postumus zum König zu machen. Bevor der Adel Wladislaw per Königswahl vor ihren Sohn setzen konnte, bemächtigte sich Elisabeth der Stephanskrone, die als heilig betrachtet wurde und deren Besitz den König von Ungarn legitimierte. Hierfür sandte sie ihre Kammerfrau Helene Kottannerin in die Plintenburg, aus der die Kottannerin die Insignie erfolgreich entführte und mit einer Schlittenfahrt über die gefrorene Donau (es war Februar) zu ihrer Königin brachte. Die Kottannerin schrieb darüber später in ihren Memoiren „Denkwürdigkeiten“. Elisabeth krönte ihren Sohn zum König von Ungarn, Kroatien und Böhmen und behielt die Stephanskrone auch, nachdem sie sie eigentlich hatte zurückgeben sollen, durch einen Betrug in ihrem Besitz. ...
 & ... Elisabeth of Luxembourg was married to the 25-year-old heir to the throne, Albert V, in 1422 at the age of 13 (they had been betrothed since she was 2 years old). After the first ten years of marriage, she gave birth to the first of their four children. Five years later, upon the death of his father, her husband became King of the Romans (Holy Roman Emperor-elect), as well as King of Hungary, Croatia, and Bohemia.
Elisabeth was five months pregnant with their fourth child when her husband died in 1439 during a military campaign against the Turks, who were invading Hungary. Despite political pressure from the nobility to marry the 15-year-old Polish king Wladyslaw III—because having a male king, regardless of his age or character, was seen as “safer” for the country in the war against the Turks—she took on the regency herself. Her goal was to secure the throne for her son, Ladislaus Postumus, as quickly as possible.
Before the nobility could elect Wladyslaw as king over her son, Elisabeth took possession of the Holy Crown of Hungary, which was regarded as sacred and essential for legitimizing the king of Hungary. To achieve this, she sent her chambermaid, Helene Kottanner, to Visegrád (Plintenburg), from where Kottanner successfully stole the crown and delivered it to her queen by sled across the frozen Danube (it was February). Kottanner later recounted this event in her memoirs, Memorabilia.
Elisabeth crowned her son as King of Hungary, Croatia, and Bohemia. Even after she was supposed to return the Holy Crown, she kept it in her possession through deceit. ...
 & ...
Entdecken Sie, wie viel eine Busfahrt von Mundo Novo nach Maracaju kostet. Verwenden Sie unsere Filter und Sortierfunktionen, um die billigsten Bus-Tickets von Mundo Novo nach Maracaju, oder Luxus-Fernbusse zu finden.
Busse, die von Mundo Novo nach Maracaju fahren, starten von der Station Terminal Rodoviaria Mundo Novo.
Ein Bus nach Maracaju wird Sie an der Station Maracaju Onibus absetzen.
Streckenplan Mundo Novo nach Maracaju
Wenn Sie im Ausland sind, sollten Sie auch etwas von dei Landessprache lernen. Auf Ihrer Busreise von Mundo Novo nach Maracaju könnte das in einer misslichen Lage sehr nützlich sein und die einheimische Bevölkerung wird sich bestimmt über Ihre Anstrengungen, eine neue Sprache zu lernen, freuen.
Freuen Sie sich bei Ihrer Busreise von Mundo Novo nach Maracaju auf einen wahren Augenschmaus mit wunderschönen Naturlandschaften und eindrucksvollen Sehenswürdigkeiten auf vielen Kilometern.
Busse haben von alle motorisierten Fortbewegungsmitteln den geringsten CO2-Ausstoss. Ein Fernbus von Mundo Novo nach Maracaju wird im Vergleich zu einem Zug nur halb so viel CO2 ausstoßen, und die Bilanz sieht im Vergleich zum Auto oder einem Flugzeug sogar noch wesentlich besser aus.
Erstellen Sie einen Soundtrack für Ihr eigenes Leben, indem Sie eine personalisierte Playlist für die Busreise erstellen. Kann es einen besseren Begleiter für Ihre Busfahrt von Mundo Novo nach Maracaju geben als Ihre Musik? ... & ... Discover how much a bus ride from Mundo Novo to Maracaju costs. Use our filters and sorting features to find the cheapest bus tickets from Mundo Novo to Maracaju, or opt for luxury coaches.
Buses traveling from Mundo Novo to Maracaju depart from the Terminal Rodoviaria Mundo Novo station.
A bus to Maracaju will drop you off at the Maracaju Onibus station.
Route Plan: Mundo Novo to Maracaju
If you are traveling abroad, it’s a good idea to learn some of the local language. On your bus journey from Mundo Novo to Maracaju, this could be very helpful in an emergency, and the locals will surely appreciate your efforts to learn a new language.
Look forward to a visual feast on your bus journey from Mundo Novo to Maracaju, with stunning natural landscapes and impressive sights stretching over many kilometers.
Of all motorized modes of transportation, buses have the lowest CO2 emissions. A coach from Mundo Novo to Maracaju will emit only half as much CO2 as a train, and the environmental impact compared to a car or airplane is even better.
Create a soundtrack for your life by making a personalized playlist for your bus journey. Could there be a better travel companion for your trip from Mundo Novo to Maracaju than your music? ... \\
\bottomrule
\end{tabularx}
\caption{Case study in German. Included are both the original German texts and their English translations.}
\label{tab:more_cases_de}
\end{table*}